%% file: icrl2024_conference.tex
\documentclass{article} 
\usepackage{iclr2024_conference,times}

\input{math_commands.tex}

\usepackage{hyperref}
\usepackage{url}
\usepackage{graphicx}
\usepackage{amsthm}
\usepackage{todonotes}
\usepackage{tcolorbox}

\newtheorem{eff_theory}{Theorem}

\newtheorem*{conjecture}{Conjecture}

\title{A Resource Model for Neural Scaling Laws}

\author{Jinyeop Song\textsuperscript{1,}\thanks{Equal contributions},\phantom{h}Ziming Liu\textsuperscript{1,2,}\footnotemark[1],\phantom{h}Max Tegmark\textsuperscript{1,2} \& Jeff Gore \textsuperscript{1} \\
\textsuperscript{1}Department of Physics, MIT, Cambridge, MA, USA\\
\textsuperscript{2}IAIFI, Cambridge, MA 02134, USA \\
\texttt{\{yeopjin, zmliu, tegmark, gore\}@mit.edu}
}

\iclrfinalcopy 

\begin{document}

\maketitle

\begin{abstract}
Neural scaling laws characterize how model performance improves as the model size scales up. Inspired by empirical observations, we introduce a \textit{resource} model of neural scaling: A task is usually composite hence can be decomposed into many subtasks, which ``compete" for \textit{resources} (measured by the number of neurons allocated to subtasks). On toy problems, we empirically find that: (1) The loss of a subtask is inversely proportional to its allocated neurons. (2) When multiple subtasks are present in a composite task, the resources acquired by each subtask uniformly grow as models get larger, keeping the ratios of acquired resources constants. We hypothesize these findings to be generally true and build a model to predict neural scaling laws for general composite tasks, which successfully replicates the neural scaling law of \textit{Chinchilla} models reported in~\citep{hoffmann2022training}. We believe that the notion of resource used in this paper will be a useful tool for characterizing and diagnosing neural networks. 

\end{abstract}

\section{Introduction}

Neural scaling laws (NSL) refer to empirically observed power law relations in large language model between the loss and model size, dataset size, and compute cost ~\citep{kaplan2020scaling,henighan2020scaling,gordon2021data,hestness2017deep,sharma2020neural,bahri2021explaining}. Several theories have been proposed to explain NSL, including the data manifold dimension ~\citep{kaplan2020scaling,sharma2020neural}, quantization of a task~\citep{michaud2023quantization}, random feature models~\citep{bahri2021explaining}, and lottery ticket ensembling ~\citep{liu2023neural}. However, the real mechanism of the neural scaling law remains unclear. The difficulty in verifying these theories stem from complex nature of both language tasks and neural networks.

Recently, there are attempts to mechanistically interpret state-of-art artificial intelligence models to elucidate their structures and functions. One key finding in this direction is the identification of modules or subnetworks~\citep{conmy2023automated, michaud2023quantization,chen2023skill,gokhale2023semantic}. Since a complicated task (e.g., language modelling) involves many subtasks, a neural network is expected to be disentangled into subnetworks or modules; each module is responsible for a corresponding subtask. 
Examples of modules include induction heads~\citep{olsson2022incontext}, shortcut for sequential operations ~\citep{liu2023transformers}, and learned algorithms for mathematical operations ~\citep{zhong2023clock, nanda2023progress, gromov2023grokking}. Throughout this paper, we will call such a subnetwork a \textit{module}~\footnote{People interchangeably use \textit{circuit}~\citep{olah2020zoom} or \textit{motif}~\citep{alon2019introduction} depending on contexts.}.

While people have discovered the existence of modules in neural networks, we are interested in the \textit{resource} perspective of these modules, i.e., how many neurons are allocated to modules? The idea is that neurons are the resource, serving as building blocks for forming any module in a network. Since a composite task involves many subtasks, neural networks must  allocate reasonable resources to each of the subtasks according to their importance. Describing the picture in anthropomorphic terms: Training of models is analogous to evolution of life. A species (in this analogy, a model) is evolved to optimally use limited nutrients (neurons) to build organs (modules for subtasks) that can best fit its habitat.



In our study, we conduct a series of toy experiments to investigate how neural networks allocate resources to single/multiple subtasks, in the presence of pressure towards sparsity. We will use \textit{the number of allocated neurons} as a metric to quantify the resources assigned to each subtask, and use these two terms interchangeably throughout the paper. Below are key findings of the paper:

\begin{figure}[t]
\begin{center}
\includegraphics[width=10cm]{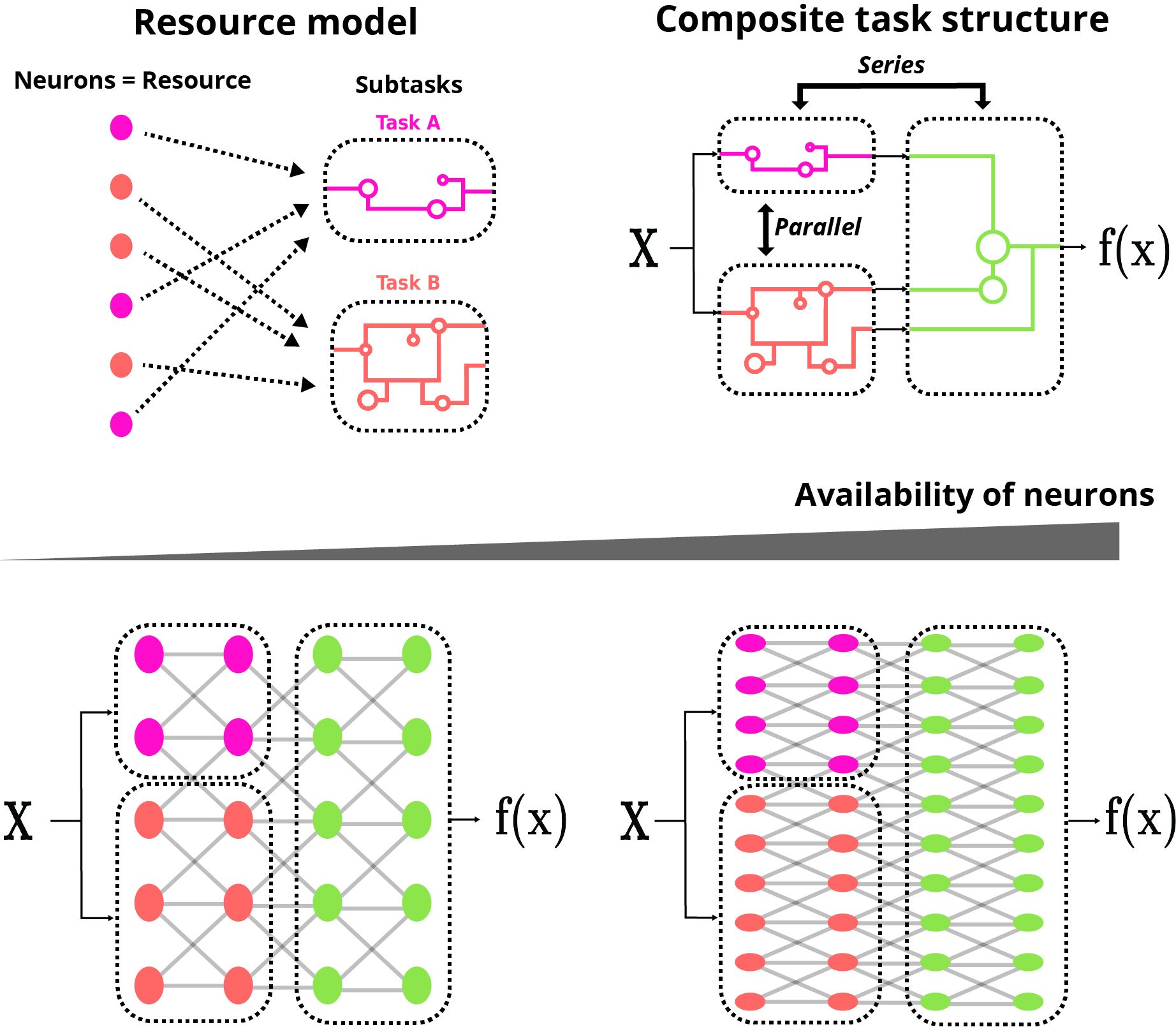}
\end{center}
\label{fig1}
\caption{Overview of the \textit{resource model}. (Top left) Neurons in a neural network play the role of resources, while (sub)tasks are consumers competing for these resources. (Top right) An example of a composite task consisting of subtasks combined in parallel and in series. (Bottom) The homogeneous growth hypothesis: when the network grows wider, each task will acquire more resources (allocated neurons), while the ratios of their acquired resources are kept constant.}
\end{figure}

\textbf{$\bullet$ For a single task: More resource allocations lead to lower losses as a power law (Section 2)}: We observe that networks can adjust the number of neurons for a module based on the importance of a subtask relative to the fixed sparsity regularization. Empirically, the relationship between the loss of a corresponding subtask (mean squared error for regression tasks or cross-entropy for classification tasks) and the number of allocated neurons \(N_{\text{subtask}}\) for the subtask is found to be roughly inversely proportional, i.e., $l_{\text{subtask}} \propto {N^{-1}_{\text{subtask}}}$.

\textbf{$\bullet$ For multiple subtasks: Homogeneous growing of allocated resources (Section 3)}:  We empirically observe that the number of neurons allocated to each subtask increases at a constant ratio as the total number of available neurons grows. Based on this observation, we propose the following conjecture: the trained states of artificial neural networks preserve constant ratios in the allocation of neurons across different subtasks. This conjecture might be equivalent to the optimal allocation of neurons subject to the finite model capacity, specifically the number of neurons in each layer. 
Combining the single task scaling law and the homogeneous growth conjecture, we can derive (and indeed observe) that the scaling law of the overall composite task can emerge, i.e., the total loss \(\ell_{\text{total}}\) is inversely proportional to the total number of neurons \(N_{\text{total}}\), i.e., \(\ell_{\text{total}} \propto {N^{-1}_{\text{total}}}\).

\textbf{$\bullet$ The implication for neural scaling laws in large language models (Section 4)}:  For a finite-width neural network trained for complex tasks which contains many subtasks, the optimal loss scales with the total resource is $\ell\propto N^{-1}$. Under a few reasonable assumptions we have $N_p\propto N^3$, so $\ell\propto N_p^{-1/3}$ ($N_p$ is the number of model parameters), agreeing with the compute-optimal scaling law of \textit{Chinchilla} models where they observed $\ell\propto N_p^{-0.34}$~\citep{hoffmann2022training}.

\section{Resource model for single task}

\begin{figure}[t]
\begin{center}
\includegraphics[width=14cm]{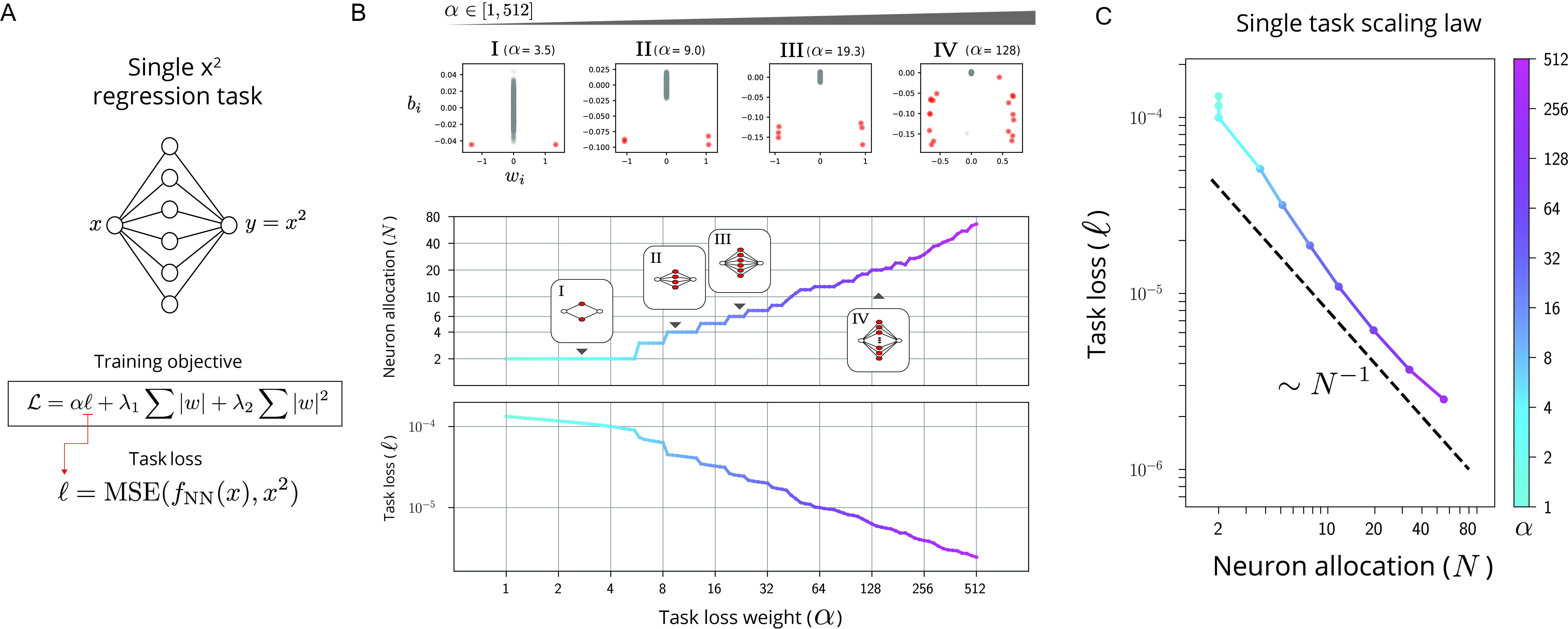}
\end{center}
\caption{(A) Toy experiment : a single $x^2$ regression task experiment. (B)(top) The weight and bias map of fixed seeds are plotted for four representative values of $\alpha=3.5, 9.0, 19.3, 128$. Dots which have nonzero weights are classified as allocated to $x^2$ module and colored as red.  Number of allocated neurons $N$ and task loss $\ell$ are plotted relative $\alpha$ (bottom). (C) The $N^{-1}$ scaling between number of allocated neurons and task loss.
\label{fig2}
}
\end{figure}

This section investigates how more resources can lead to lower losses of a subtask, predictably as a power law with a scaling exponent of $-1$. We have the following hypothesis and use a toy experiment to support the hypothesis.

\begin{center}
\begin{tcolorbox}[colframe=black, boxrule=1pt, colback=white, width=0.80\textwidth]
{\bf Resource model hypothesis 1 (Scaling law of a single subtask)}: For a single subtask, if a network allocates \(N\) neurons for that subtask, the network can achieve a loss that scales as \(\ell \sim {N^{-1}} \).
\end{tcolorbox}
\end{center}

{\bf Experimental setup} Let us consider a simple regression task in figure ~\ref{fig2}-A.
The fully connected network has a single hidden layer with 1000 neurons, with a scalar input $x \sim \text{Uniform}[a, b]$ and aims to predict $y=x^2$. The network is parametrized as

\[f_{NN}(x) = \sum_{i=1}^{N} v_i \sigma(w_i x + b_i) + c\]

where \(\sigma\) is the SiLU activation function, and $\text{\boldmath$\theta$}=\{v_i,w_i;i=1,2,\cdots,N\}$ denote all weight parameters. 
 The task loss $l$ of our experiment is MSE of the network output $\hat{y}=f_{NN}(x)$ compared to the target function $y=x^2$,

\[l=\langle (\hat{y}-y)^2\rangle\] 

where $\langle\cdot\rangle$ is averaging over data samples. We also add penalty pressure to encourage sparsity and small norms, so the total training objective is
\[L=\alpha l+\lambda_1\sum\|\theta_w\| + \lambda_2\sum{\|\theta_w}\|^2\]
where $\alpha$ is a hyperparameter emphasizing the relative importance of task loss compared to regularization. $\sum\|\theta_w\|$ and $\sum{\|\theta_w}\|^2$ are L1 and L2 norms of weights, respectively. We sweep $\alpha\in [1,512]$.  
  We fix $\lambda_{1} = 0.0001$ and $\lambda_{2} = 0.0005$. Such dual regularization approach aims to reduce training variability and foster a clear module structure. The Adam optimizer is used to train the neural network with total epochs set to 100000.  The learning rate is initially started at 0.01 and reduced by a factor of 0.3 every 20000 epochs. The batch size is set to 500.

Our toy experiment is designed to answer two questions regarding resources (the number of allocated neurons): (1) How many neurons are used to do the task (while others are useless)? (2) How do more resources contribute to reducing task losses? Hypothetically the balance between the task loss and sparsity pressure determine the amount of resources used to perform the task, which we verify below.

\textbf{Results} We train a network with a fixed seed, varying $\alpha\in[1,512]$ and plot the number of allocated neurons, and task loss in figure \ref{fig2}-B. Allocated neurons were identified by a nonzero $(>10^{-3})$ variance of activation over various inputs $x$ and annotated as red dots. 
we are able to observe that high $\alpha$ indeed promotes formation of more allocated neurons. Next, We look into how the task loss decreases as the number of allocated neurons increases. Notably, we observe a steep decline in task loss exactly at the same time as the number of allocated neurons suddenly jumps. This pattern indicates that the increase in resources might be crucial for reducing task loss in this model.

We train networks over many random seeds ($n_{\rm seeds}=16$ for each $\alpha$ values) and plot the relationship of the number of allocated neurons $N$ and the task loss is shown in figure \ref{fig2}-C. Notably, we observe a $N^{-1}$ scaling relation between task losses and the number of allocated neurons. The summary of the toy experiment is: the task loss is roughly $N^{-1}$, where $N$ is the number of allocated neurons. 

{\bf Remark} (Scaling laws of various regression tasks) : We further investigated whether $\ell\propto N^{-1}$ scaling law holds across regression tasks with various target functions(see Appendix \ref{appendixA} for details). We observe that majority of chosen functions roughly scales with exponent of -1, but with a case of exception where loss scales faster than $-1$ at the initial phase.

{\bf Remark} (Mechanism behind the $\ell\propto N^{-1}$ scaling): One might question the underlying mechanism for the observed $\ell\propto N^{-1}$ scaling law. In this regard, we investigated if ensembling mechanism, which indicates ensembling independent and orthogonal outputs can reduce an error, could account for the observed scaling relation(see Figure \ref{fig:supp_2} in Appendix for details). However, our analysis revealed that the ensembling does not occur in the current experiment.

\section{Neuron redundancy in Composite tasks }

\begin{figure}[t]
\begin{center}
\includegraphics[width=9cm]{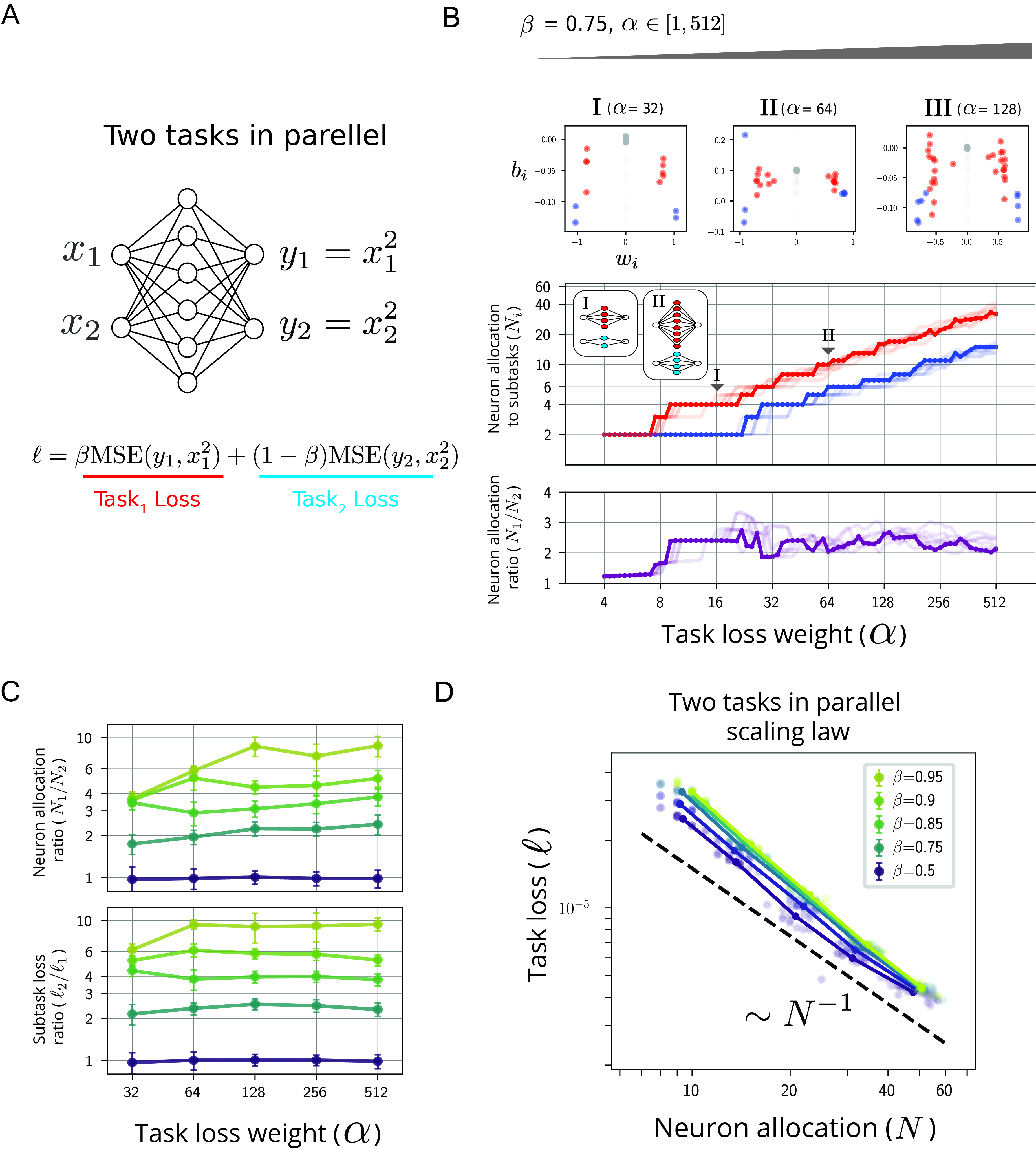}
\end{center}
\caption{(A) A neural network with one hidden layer is tasked to perform two independent squared regression tasks in parallel. We vary hyperparameters $\alpha$ for overall task intensity and $\beta$ for relative weights of two tasks. We annotated neurons allocated for the first and the second tasks red and blue colors, respectively. (B) Resources of the first and second tasks increase together as $\alpha$ increases. Surprisingly, we observe the constant ratios of allocated neurons across a range of $\alpha$, which serve as the basis for the \textit{homogeneous growth of resource allocation} hypothesis. (C) The ratios of MSE and ratios of allocated neurons are insensitive to $\alpha$ but depend heavily on $\beta$. (D) We observe the emergence of $N^{-1}$ scaling for the composite task of two tasks combined in parallel.}
\end{figure}
\label{fig3}

Practical tasks are often far more complex than $x^2$ regression. It is common for certain task to be a (possibly complex) composition of multiple subtasks. There are two types of basic compositions: composition in parallel or composition in series. An example of composition in parallel is a computer vision model tasked with inferring the overall context of an image. The model is expected to detect various classes of objects together, such as cars, animals and human. 
An example of composition in series is a model expected to build the flow of logical reasoning: the model first recognizes the objects, an then interprets the relationships between objects, and finally deduces the overall context.

Our question is: when a task is a composition of multiple subtasks, how does a neural network allocate its resources to accommodate all subtasks? In particular, we are curious how do resource allocations change when the neural network is scaled up? We have the following hypothesis:

\begin{center}
\begin{tcolorbox}[colframe=black, boxrule=1pt, colback=white, width=0.80\textwidth]
{\bf Resource model hypothesis 2 (\textit{Homogeneous growth of neuron allocation})}: 
If a neural network with  $N$ neurons allocates resources $N_i$ to subtask $i$, then if the network size is scaled up (along width) such that $N\to a N$, all the resources are scaled up by the same factor $a$, i.e., $N_i\to aN_i$.
\end{tcolorbox}
\end{center}

We will use toy experiments below to show that the \textit{homogeneous growth of resource allocation} is true in two composite scenarios: tasks in parallel and tasks in series.

\subsection{Two tasks in parallel}
\textbf{Experiment setup} We start with the two tasks connected in parallel as in figure \ref{fig3}-B. The network has a single hidden layer with 1000 neurons, with a two dimensional vector $(x_1,x_2)$ s.t $x_1,x_2 \sim \text{Uniform}[-1, 1]^2$ and aims to predict $(y_1, y_2)=(x_1^2, x_2^2)$. Note that the task is basically the parallel composition of two $x^2$ regression tasks. We set the different relative weights for the first and second task, so the task loss $l$ from the network output $f_{NN}(x_1,x_2)=(\hat{y_1},\hat{y_2})$ is:

\[
l=\beta \langle(\hat{y_1} -y_1)^2\rangle + (1-\beta) \langle ( \hat{y_2}-y_2 )^2\rangle 
\] 

where $\langle\cdot\rangle$ means averaging over data samples, $\beta$ is relative intensity of subtasks sampled within a range $[0,1]$. The training loss function is defined as

\[L=\alpha l + \lambda_1\sum\|\theta_w\| + \lambda_2\sum{\|\theta_w}\|^2 \]

We use the same experimental conditions as in our previous toy experiment.

\textbf{Results} We train a network with fixed seeds and $\beta=0.75$, while varying $\alpha$ within the range [1,512]. We then plot the weight and bias information of each neurons toward two inputs. Each neuron was allocated to either the first or the second regression task, characterized by their nonzero weights in the first layer connected to the input neurons. Notably, we did not observe the occurrence of superposition in the experiments, i.e., a single neuron is either allocated completely  to $x$ or completely to $y$ or useless~\footnote{Superposition, the phenomenon where a neuron represents two or more distinct concepts, is often observed in practical models (\cite{elhage2022superposition}). However, we did not observe superposition in our experiments. This may be attributed to the model size being sufficiently large relative to the complexity of the given tasks, differing from the typical conditions for superposition occurrence, where the representation dimension exceeds the model size (\cite{elhage2022superposition}).}. 
We plot the ratio of number of neurons allocated to the first task respect to that of the second task in figure \ref{fig3}-B. We observe that this ratio remains constant at 2.35 $\pm$ 0.24 ($n_{seeds}=8$) across a wide range of $\alpha$ values. Based on these observations, we hypothesize that the ratio of allocation for each task is determined by the relative task intensity rather than the absolute intensity. To investigate this hypothesis, we now trained a network many times with various seeds($n_{seeds}=16$), varying $\alpha=32,64,128,256,512$ and $\beta=0.5, 0.75, 0.85, 0.9, 0.95$. We observe that both the ratio of number of allocated neurons and MSE of the first task in respect to the second tasks are maintained constant \footnote{At low alpha discrete nature of number of neurons are making the ratio sudden jumps to constant value. This phenomena is obvious particularly when alpha is low and beta is high, so that the second task is extremely low and corresponding neurons are kept 2 or 4.}. This is consistent with the homogeneous growth hypothesis.

The \textit{resource model} enables to establish the scaling relation of tasks in parallel. According to {\bf Resource model hypothesis 1}, the loss associated to each subtask inversely scales with the number of allocated neurons, i.e., ${\rm MSE}_i \sim {N_i}^{-1}$. If it is assumed that the proportion of neurons allocated to each subtask remains constant as \(N\), the total number of allocated neurons (\(N = N_1 + N_2\)), varies, then the normalized composite task loss is given by: 
\[
l_{global} = \beta {\rm MSE}_1 + (1-\beta) {\rm MSE}_2 \sim \frac{\beta}{N_1} + \frac{1-\beta}{N_2} \sim \frac{1}{N}
\]

This scaling between task loss and the total allocated neuron count is consistent with our observation for various values of $\beta = 0.5, 0.75, 0.8, 0.9, 0.95$, as shown in Figure 1-D. Based on the {\bf Resource model hypothesis 1} and {\bf 2}, combined with the linear additivity of losses, we have

\begin{eff_theory}[Scaling of Tasks Combined in Parallel]  
Consider a set of target functions \( f_i: x \rightarrow f_i(x) \) for \( i=1,2,\ldots,Q \), and corresponding number of allocated neurons \(N_i\). Consider a composite task, which is a linear and parallel combination of these regression tasks i.e composite task loss is \(l = \sum_{i \in Q}\alpha_i {\rm MSE}_i\). Assuming a fixed relative ratio of resources for each subtask (the homogeneous growth hypothesis), the composite loss scales as \(\sim N_{\text{total}}^{-1}\), where \(N_{\text{total}} = \sum_i N_i \) is the total number of neurons allocated to the composite task.
\end{eff_theory}

{\bf Remark} (Origin of \textit{homogeneous growth of resource allocation}) \textit{Homogeneous growth of resource allocation} may naturally arise from the optimal solution to resource allocation problem. We explore this possibility through a constraint-optimization problem, detailed in the Appendix \ref{appendixC}.



\subsection{composition of tasks in series}

Now we consider tasks arranged in series. A simple example is the regression task of a composite function. We aim to perform regression on the function \( F(x_1, x_2, x_3, x_4) = \sqrt{(x_1 - x_2)^2 + (x_3 - x_4)^2} \). This function is a composition of two distinct functions \( f \circ g \), where \( f(x) = \sqrt{x} \) and \( g(x_1, x_2, x_3, x_4) = (x_1 - x_2)^2 + (x_3 - x_4)^2 \).

\textbf{Toy experiment setup} Consider a network with four hidden layers, each consisting of 30 neurons. The network processes a four-dimensional input vector \( (x_1, x_2, x_3, x_4) \), where each \( x_i \) is independently sampled from a Uniform distribution $U[-1, 1]$. The task loss $l$ from the network output $f_{NN}(x_1, x_2, x_3, x_4)=(\hat{y})$ is
\[
l=\langle(\hat{y} - \sqrt{(x_1 - x_2)^2 + (x_3 - x_4)^2})^2 \rangle
\] 
where $\langle\cdot\rangle$ denotes averaging over data samples. The training objective is defined as

\[
L = \alpha l + \lambda_1\sum\|\theta_w\| + \lambda_2\sum{\|\theta_w}\|^2
\]

We use the same experimental conditions as in our previous experiments, except that we sample \( \alpha \) from the range $[20, 10240]$ and set the number of training epochs to \( 3 \times 10^5 \). This adjustment in the experimental setup is intended to speed up convergence.

\begin{figure}[t]
\begin{center}
\includegraphics[width=14cm]{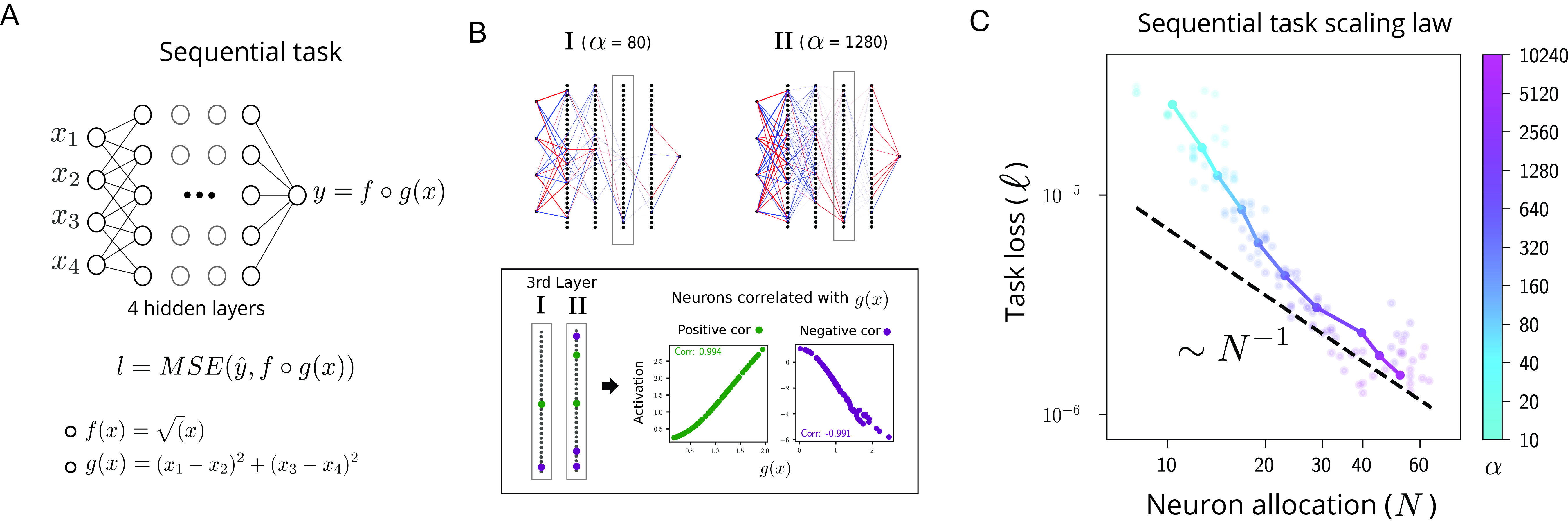}
\end{center}
\caption{ Composition of tasks in series (A) A neural network with four hidden layer is tasked to perform the regression of composite function of $f$ and $g$. We vary hyperparameters $\alpha$ for overall task intensity. (B) (top) Weight structures of trained model are visualized for $\alpha=80, 1280$. (bottom) Neurons in the third layers of both model exhibit high correlation with the $g(x)$, green for postiive and purple for negative correlation. We found the evidence that the first three layers perform $g(x)$ and the last layer perform $f(x)$. (C) We observe the $N^{-1}$ scaling for the composite task loss of two tasks combined in series.}
\label{fig4}
\
\end{figure}

\textbf{Results} In our experiment, we train a model with fixed seeds while varying the parameter $\alpha$. As $\alpha$ increases, the network's density also increases, a trend depicted in Figure 4-B. Notably, in the network's third layer, we identify a specific set of neurons. These neurons approximate the function $(x_1 - x_2)^2 + (x_3 - x_4)^2$. Within this set, we consistently observed two distinct classes of neurons: one class positively correlated (marked in green) and the other negatively correlated (marked in purple). This pattern indicates that the model processes the $f \circ g$ computation in a sequential manner. We postulated that the first three layers seem to represent the $module$ performing $g$, while the final layer corresponds to the $module$ performing $f$.


Next, we investigate the relationship between the number of allocated neurons and task loss, varying $\alpha$ in Figure 4-C. Initially, the task loss exhibits a rapid scaling, but with increasing \(\alpha\), it adheres to an \(N^{-1}\) scaling.  Notably, the transition to the scaling of $N^{-1}$ is consistent with Figure 4-B, where the fraction of nonzero neuronal redundancy get less than 50\% around $\alpha = 640$. We propose that our dataset within the chosen \(\alpha\) regime reveals two distinct phases. In the first phase, where loss steeply decrease faster than $N^{-1}$, possibly indicating a qualitative change in neuronal strategies, such as the introduction of neurons with unprecedented functions as \(\alpha\) increases. In the second regime, where loss scales as $N^{-1}$, redundant neurons are introduced to reduce task loss. Investigating this hypothesis further could elucidate the conditions under which \(N^{-1}\) scaling emerges across various regression tasks.

Furthermore, we checked if the \textit{resource model} can  establish the observed scaling relation of tasks in series. First, we demonstrated that, under certain assumptions (errors of subtasks should be orthogonal and independent)\footnote{However, it remains to be uncertain whether these prerequisites apply to the composite function in the toy experiment.}, composite task loss of sequential tasks is linearly seperable to sum of subtask losses(see Appendix \ref{appendixC} for details). Next, according to {\bf Resource model hypothesis 1}, the loss associated to each subtask inversely scales with the number of allocated neurons, i.e., ${\rm MSE}_f \sim {N_f}^{-1}$ and  ${\rm MSE}_g \sim {N_g}^{-1}$. It allows to establish the scaling law as follows : 

\[
l = \left\|f(g(x)) - f(G(x))\right\|^2 + \left\|f(G(x)) - F(G(x))\right\|^2 \simeq \frac{A}{N_g} + \frac{B}{N_f},
\]

where \(A\) and \(B\) are constants. 

Furthermore, we examined the validity of the Homogeneous Growth of hypothesis for experiments involving tasks in series. We analyzed how the number of live neurons in each layer varies with respect to \(\alpha\), as depicted in Figure~\ref{fig:supp_3} in Appendix. We indeed observe a coherent increase in the number of active neurons across all layers. Therefore, under the assumption of {\bf Resource model hypothesis 2}, the loss scales with the total number of allocated neurons

\[
l \simeq \frac{A}{N_g} + \frac{B}{N_f} \sim \frac{1}{N}.
\]

Summarizing the above argument, we have established the following theorem : 
\begin{eff_theory}[scaling of tasks combined in series]: Consider a set of target functions $f_i: x \rightarrow f_i(x) \ (i=1,2,\ldots,Q)$. Consider a composite task, which is a serial combination of such regression tasks i.e. \(l = MSE(\hat{y}, (f_1 \circ f_2 \circ f_3 \circ \ldots \circ f_Q)(x))\).  Assuming a fixed relative ratio of resources for each subtask  (homogeneous growth of resource allocation) and a linear seperability of the composite task loss function, then the total loss scales $\sim N_{\text{total}}^{-1}$, where \(N_{\text{total}} = \sum_i N_i \) is the total number of neurons allocated to the composite task.
\end{eff_theory}

\subsection{Conjecture for general composite tasks}

We have shown that, for simple composition cases (in parallel and in series), the loss function is linearly separable for subtasks, i.e., the loss function can be written as the sum of functions of resources for each subtask. We conjecture this linear separability of loss can apply to general composite tasks.

\begin{center}
\begin{tcolorbox}[colframe=black, boxrule=1pt, colback=white, width=0.80\textwidth]
{\bf Resource model hypothesis 3 (Linear additivity of subtasks losses)}: 
The loss of a composite task can be decomposed as a linear summation of the losses of its subtasks. 
\end{tcolorbox}
\end{center}

Combining all three hypothesis, we can derive that 

\begin{center}
\begin{tcolorbox}[colframe=black, boxrule=1pt, colback=white, width=0.80\textwidth]
{\bf The scaling law of general composite tasks}: 
Given three hypothesis of the resource model being true, we can derive that the loss $\ell$ of a general composite task scales as $\ell\propto N^{-1}$, where $N$ is the number of neurons contributing to the overall task.
\end{tcolorbox}
\end{center}


\section{Emergent Neural Scaling of model size}

In previous sections, we demonstrated that $\ell\propto N^{-1}$, i.e., the loss $\ell$ is a scaling law against the number of allocated neurons $N$. 
Can we use this result as a starting point to derive the more useful neural scaling laws of large language models (LLMs)? Notice that there are two major gaps between our resource model and the LLM setup: (1) In LLM, the number of parameters $N_p$ is measured against instead of the number of allocated neurons $N$; (2) In LLM, the depth is usually scaled with width, while our resource model (hypothesis 2) assumes only scaling up width. Below we aim to address these two gaps and derive a meaningful scaling law for LLMs.

In LLMs the empirical neural scaling law we care about is $\ell\propto N_p^{-\alpha}$ where $N_p$ is the number of model parameters. How is $N_p$ related to $N$? We aim to show that $N_p\propto N^3$ and finally derive $\alpha=1/3$, with a number of reasonable assumptions\footnote{we denote $N$ as effectively represents the overall resources (number of allocated neurons) of various subtasks. The homogeneous growing of resource conjecture enables to regard the resources of various tasks as a single quantity.}:

{\bf Assumption 1:} The number of modules is proportional to the network width, i.e., $N\propto N_{\rm width}$. This is aligned with the homogeneous growth conjecture. Redundant representations have been empirically observed in~\cite{liu2023neural,doimo2021redundant}.

{\bf Assumption 2:} The depth is scaled as the same factor as the width, i.e., $N_{\rm depth}\propto N_{\rm width}$. The theoretical derivation of the optimal depth-width ratio~\citep{levine2020depth} has been the foundation for best practices in LLM where the depth and width of the network are scaled by the same factor, e.g., in~\cite{hoffmann2022training}. 

{\bf Assumption 3:} The number of parameters can be roughly estimated as $N=N_{\rm width}^2N_{\rm depth}$. Most neurons of the neural network are in MLPs weights, while the number of other parameters are nearly negligible. The number of MLP weight parameters is simply $N_{\rm width}^2N_{\rm depth}$. 

Combining all three assumptions, we have
\begin{equation}
    N_p = N_{\rm width}^2 N_{\rm depth}\propto N_{\rm width}^3 \propto N^3,
\end{equation}
together with $\ell\propto N^{-1}$ immediately implies that $\ell\propto N_p^{-1/3}$. Pleasantly, this agrees with the empirical scaling law of Chinchilla models~\citep{hoffmann2022training} where they observed $\ell\propto N_p^{-0.34}$. Since our theory has no hyperparameter at all hence does not have the ability to ``fine-tune" to any empirical data, it is indeed surprising that our theory can predict~\citep{hoffmann2022training} amazingly well. However, it could be a pure coincidence. 

Also, we would like to point out something counterintuitive when assumption 1 and 2 are considered together, which assumes that $N$ is independent of $N_{depth}$. We conjecture: (1) there exists a critical depth such that above the critical depth, further increasing depth does not help; (2) current LLMs are beyond the critical depth. If these conjectures are indeed true, they would then give a testable hypothesis that the better way to scale up current LLMs is by scaling up the width but keeping the depth constant, which will give a faster scaling $\ell\propto N_p^{-\frac{1}{2}}$.


\section{Related works}

\textbf{Neural Scaling laws (NSL)} is the phenomenon where model test losses decreases as power laws as model size, compute, or data scale up~\citep{kaplan2020scaling,henighan2020scaling,gordon2021data,hestness2017deep,sharma2020neural,bahri2021explaining}. The origin of neural scaling laws is still debatable, but there are a few plausible theories from data manifold dimension~\citep{kaplan2020scaling,sharma2020neural}, ``quantization" of a task into subtasks~\citep{michaud2023quantization}, or random feature models~\citep{bahri2021explaining}. Our paper proposes yet another neural scaling mechanism from module multiplicity.


\textbf{Neuronal redundancy} is refering to phenomenon that there are more neurons in brains than are strictly necessary for basic functioning.
Recently, it is actively explored the similar concept of neuronal redundancy in artificial neural network in terms of overparameterization~\cite{pmlr-v97-allen-zhu19a} and their roles~\cite{pmlr-v139-liu21y, zou2019improved}. 

\textbf{Emergence} refers to the phenomenon where a model has abrupt improvements in performance when its size is scaled up [13], although this can depend on specific metrics [14]. Our discussion on ensembling versus synergy may further provide a useful tool to categorize types of emergence: trivial (pure ensembling) or non-trivial (with synergy).


\textbf{Mechanistic interpretability} is a research field aiming to mechanistically understand internal computations and representations of neural networks~\citep{nanda2023progress,wang2022interpretability,zhong2023clock}. Works in mechanistic interpretability focus on identifying semantically meaningful neurons~\citep{gurnee2023finding}, heads~\citep{olsson2022context} or circuits~\citep{conmy2023towards}. Our work is an example of module (circuit) level analysis, suggesting that the apparent complexity of neural networks may come from many copies of the same module.

\textbf{Optimization} in artificial intelligence often involves heuristic and empirical methods. Our resource model suggests that optimal allocation of neurons to each subtasks are crucial for enhancing performance. In this sense, our model validates the efficacy of Normalization \citep{DBLP:journals/corr/IoffeS15, ba2016layer} and dropout \citep{JMLR:v15:srivastava14a}, which encourage the comprehensive utilization of neurons.

\section{Conclusion}

In this work, we introduced a novel \textit{resource model}:
When a task is a (possibly complex) composition of subtasks, each subtask corresponds to a module in the neural network consuming certain amount of resources. We found the resource model can explain neural scaling laws in our well-controlled toy setups: a single task, tasks composed in parallel and tasks composed in series. Our empirical results show good evidence for the effectiveness of the resource model, leading to many phenomenon which could be interesting on their own sake, including lack of superposition and neuron redundancy. Although this resource model is humbly framed as mainly hypothesis, the agreement from toy experiments give the hope that this simple yet effective framework can be extended to realistic models, e.g., large language models.

\section*{Acknowledgement}

We would like to thank Yizhou Liu for helpful discussion. J.S. and J.G. acknowledges support from the Alfred P. Sloan Foundation and Schmidt Polymath Award. Z.L. and M.T. are supported by the Rothberg Family Fund for Cognitive Science and IAIFI through NSF grant PHY-2019786.

\bibliography{iclr2024_conference}
\bibliographystyle{iclr2024_conference}

\newpage

\appendix

\section{Scaling law across various regression tasks}
\label{appendixA}

\begin{figure}[h]
\begin{center}
\includegraphics[width=14cm]{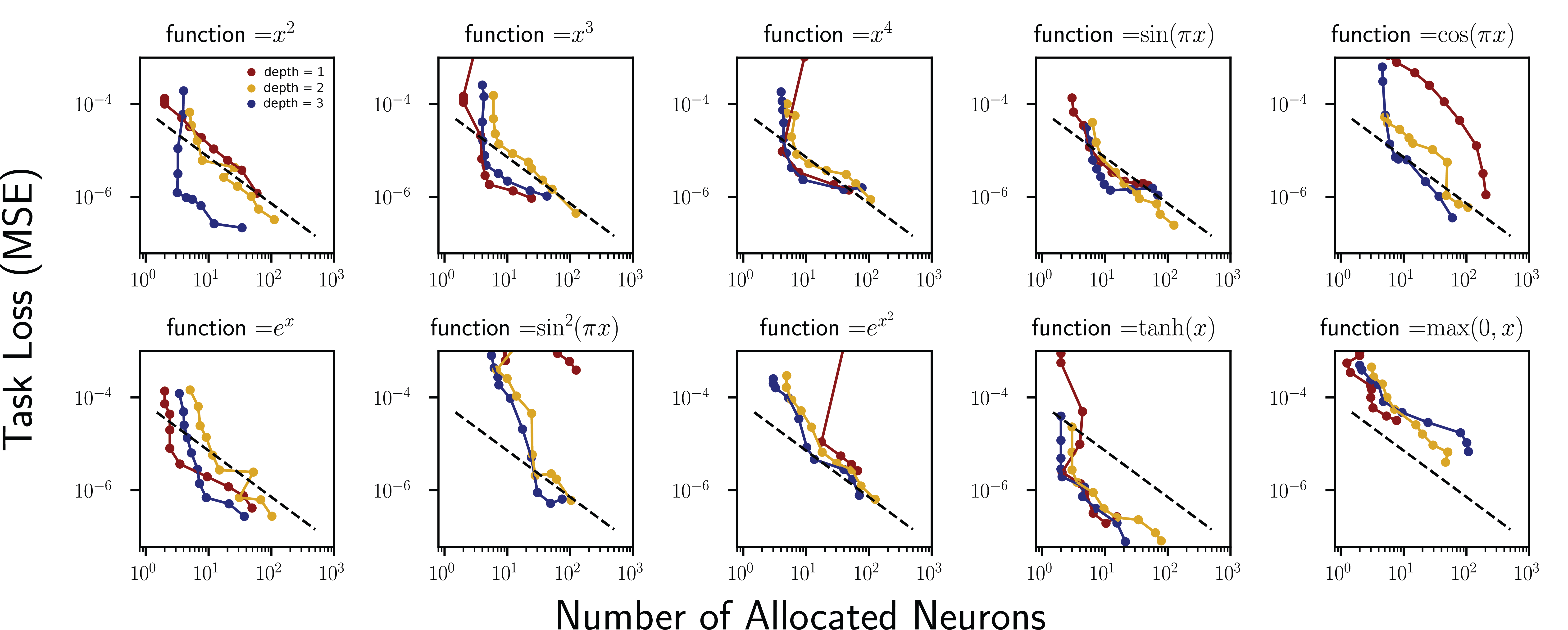}
\end{center}
\caption{\label{fig:supp_1}Scaling relationship between the number of allocated neurons and task loss across various regression tasks. Simulations were conducted for three different network depths, with hidden layer configurations of [1000], [200, 200], and [150, 150, 150], which are represented by the colors red, yellow, and blue, respectively.}

\end{figure}

 We investigated the that $N^{-1}$ scaling law between task loss and number of allocated neurons holds across various simple function regression tasks, as shown in\label{fig:supp_1}. Experiments were performed with identical settings in [Section 2] except changing the target functions. We also trained the model with $depth=1,2,3$ to address the possibility that some target functions might need more than a single layer to form a module. We observe that majority of functions shows -1 scaling law asymtotically, after the faster decrease phase, except for $sin^2(\pi(x))$.

\section{Discussion on mechanism of $\ell\propto N^{-1}$ scaling law}

\subsection{Ensembling mechanism}
\label{appendixB}

In this section, we describe the ensembling mechanisms that ensembling of multiple independent and identical outputs can reduce uncorrelated errors. lets focus on regerssion tasks, while classification tasks are described in the later section. In order for network to have $n$ multiple independent and identical outputs, there should be $n$ seperated subgraphs of the similar topology but with slightly different parameters in the network. We will call such subgraphs as regressors. The total number of allocated neurons are proportional to number of regressors, i.e $N = nk$, where $k$ is a number of neurons to form a regressor.

The following Lemma suggests a regressor has zero mean of error.

\textbf{Lemma }(Zero Mean Error) Consider a simple regression task $x \to f(x)$ and corresponding regressor that performs $F:x \to F(x)$ to minimize $l=E_{x \in U}\|f(x)-F(x)\|^2$, where $U$ is input distribution. Then, the expectation of error is zero. 
\[E_{x \in U}[f(x)-F(x)]=0\]

\textbf{Proof} Suppose that $\overline{f} = E(f(x))$ and $\overline{F} = E(F(x))$. Then 
\begin{align*}
l &= E_{x \in U}\|f(x)-F(x)\|^2 \\
  &= E_{x \in U}\|f(x)-\overline{f}(x) + \overline{f}(x) - F(x)\|^2 \\
  &= E_{x \in U}\|f(x)-\overline{f}(x)\|^2 + E_{x \in U}\|\overline{f}-F(x)\|^2 \\
  &= E_{x \in U}\|f(x)-\overline{f}\|^2 + E_{x \in U}\|F(x)-\overline{F} + \overline{F}-F(x)\|^2 \\
  &= E_{x \in U}\|f(x)-\overline{f}\|^2 + E_{x \in U}\|F(x)-\overline{F}\|^2 + \|\overline{f}-\overline{F}\|^2
\end{align*}

The last term $E_{x \in U}\|\overline{f}-\overline{F}\|^2$ denotes the deviation of mean between of regressor output to true function. The neural network with bias terms has capability to reduce the last term to zero. Therefore, $\overline{f}=\overline{F}$ In summary, bias term is now can be ignored. 

From now on, We assume that $\overline{f}=\overline{F}=0$ without loss of generality.

\textbf{Definition : Independent and Identical Regressors} Consider a simple regression task $x \to f(x)$ and a model that have multiple regressors $M_i$ ($i\in {1,2,..N_M})$ performs $F_i:x \to F_i(x)$. The regressors  are considered to be independent and identical distributed if their error functions $e_i(x)=f(x)-F(x)$ satisfy two conditions : firstly, $E_{x \in U}[e_i(x)]=0$ and secondly, $E_{x \in U}[e_i(x) e_j(x)]=\hat{e}^2\delta_{ij}$, where $\hat{e}^2$ as mean squared error of each individual module.

\textbf{Ensembling mechanisms} Consider a regression task \( f \):\( x \to f(x) \). Assume that a model \(\mathcal{M}\) comprises \( n \) independent and identical regressors \( M_1, M_2, \ldots, M_n \), where each regressor performs a function \( F_i: x \to F_i(x) =f(x)+e_i(x)\) with a mean squared error (MSE) of \(var(e(x))=\hat{e}^2 \) in a specific layer. We further assume that error is negligible compared to the function norm i.e $\hat{e} <<\|f\|$ and error is orthogonal.  Then, the model is able to perform the regression task with MSE of $\hat{e}^2/N$ in the subsequent layer through ensembling of \( M_1, M_2, \ldots, M_n \). 

\textbf{Proof} Consider a neuron in the subsequent layer receives outputs from regressors and perform summation. These modules are associated with weights \( w_1, w_2, \ldots, w_N \), so that the activation of the neuron is \( F^*: x \mapsto F^*(x) = \sum_{i}w_iF_i(x)\) Then the neuron has the following MSE.
\begin{align*}
l & = E_{x \in U}\|f(x)-F^*(x)\|^2 \\
  & = E_{x \in U}\|f(x)-\sum_{i}w_iF_i(x)\|^2 \\  
  & = E_{x \in U}[1-\sum_{i}w_i]^2 \|f(x)|^2 + 2E_{x \in U}[1-\sum_{i}w_i f(x)+b^*][\sum_{j}e_j(x)]+ E_{x \in U}\sum_{i}\sum_{j}w_iw_j[e_i(x) e_j(x)] \\  
  & = E_{x \in U}\sum_{i=1,2,..n}\sum_{j=1,2,..n}w_iw_j[e_i(x) e_j(x)] = \sum_{i} w_i^2 e_i(x)  \geq \frac{\hat{e}^2}{n}
\end{align*}

The equality holds when $w_1=w_2=..=w_n$

\subsection{Matching Ensembling mechansim with $x^2$ regression experiment}

\begin{figure}[t]
\begin{center}
\includegraphics[width=12cm]{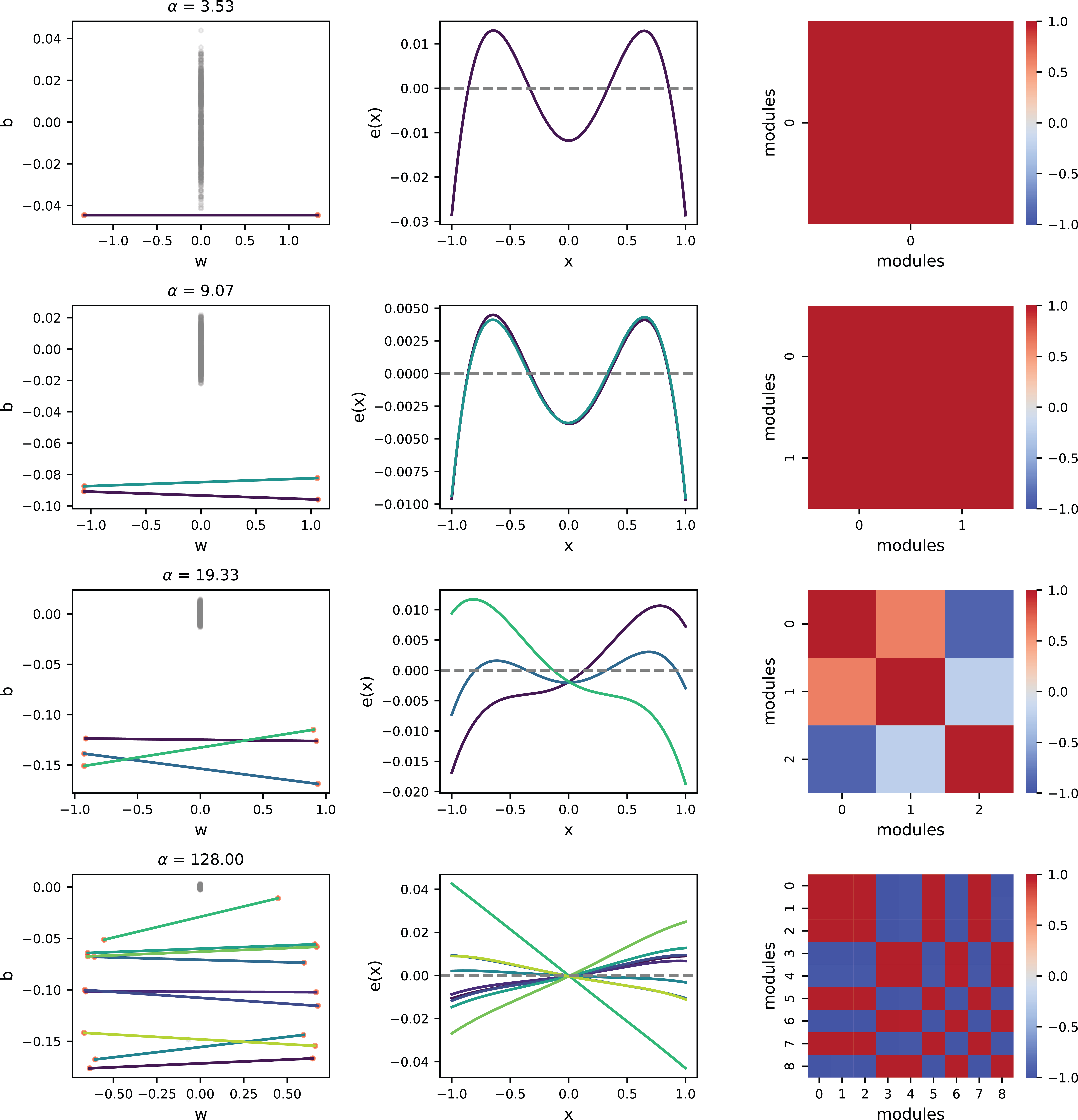}
\end{center}
\caption{Error functions of $x^2$ regressors in the toy experiment. $alpha=3.53, 9.07, 19.33, 128$ respectivley. from top to bottom (left) Plots show weight and bias of the first layer. Neurons exhibiting symmetry are identified and paired based on their relative distances when projected onto their counterparts, annotated by colored lines. (center) the error function profiles for the regressors with respect to the input $x$. (right) The correlations between error functions of each regressors are plotted}
\label{fig:supp_2}
\
\end{figure}

 We revealed that the ensembling mechanism, suggested in ~\cite{liu2023neural}, does not convincingly account for loss scaling observed in a toy experiment for single $x^2$ regression task. In the experiment, ensembling is expected to occur between outputs from several $x^2$ regressors, pairs of symmetric neurons that have the same bias and the opposite weight in the first layer ~\cite{liu2023neural}. We identified multiple $x^2$ regressors through matching pair of neurons by minimizing relative distances when projected onto their counterpart as in Figure 6(left). We ploted the error functions(center) across the input distribution and their correlations(right). Correlation values are $\pm 1$ in most cases, indicating these error function are highly correlated(anticorrelated). Therefore, we concluded that ensembling does not occur for a set of $x^2$ regressors in our single $^2$ regression toy experiment.

\section{Origin of homogeneous growth of resource allocation}
\label{appendixC}

\textbf{Plausible scenario about \textit{Homogeneous growth of resource allocation}} It is insightful to note that the \textit{Homogeneous growth of resource allocation} could be derived by assuming optimal solution of resource allocation. For instance, consider the following optimization problem in our two-task experiments:

\begin{align}
 \text{minimize} \quad & l(n_1, n_2) = \frac{a}{n_1} + \frac{b}{n_2},  \\
 \text{subject to} \quad & 2(n_1 + n_2) = N.  
\end{align}

The optimal solution results in a constant ratio between $n_1$ and $n_2$. This principle can be extended to any composite task where the loss function is \textit{linearly separable} to sum of subtasks loss. Therefore, it implies that once \textbf{Resource model hypothesis 1} and \textbf{Resource model hypothesis 3} hold, then \textbf{Resource model hypothesis 2} naturally follows assuming the optimal solution.

\section{Scaling law in composite tasks}

\label{appendixD}
\subsection{Neuron redundancy measure}

\begin{figure}[h]
\begin{center}
\includegraphics[width=14cm]{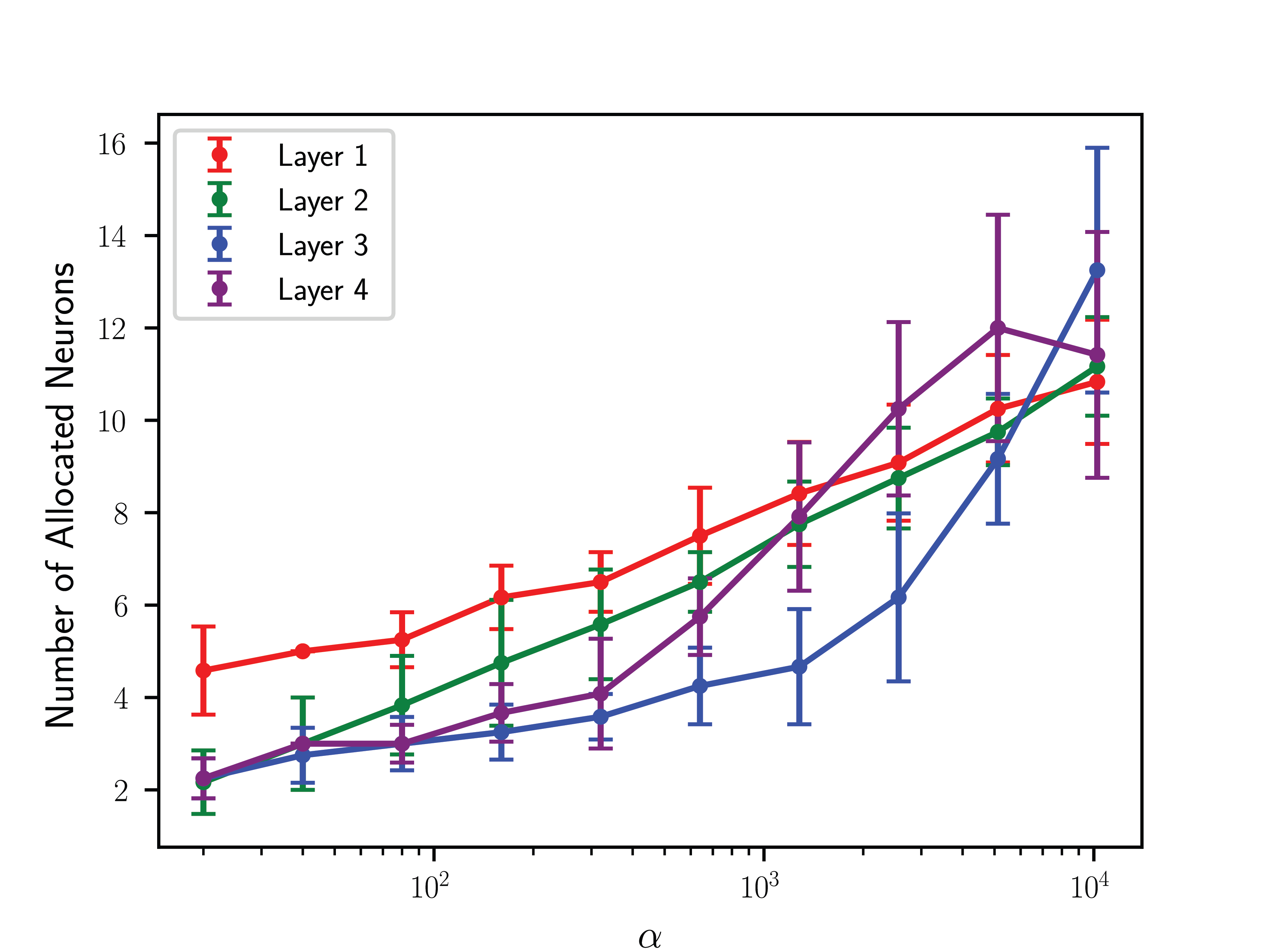}
\end{center}
\caption{Neuron redundancy, defined by a number of other neurons in a layer that has highly correlated activation patterns, are plotted as respect to $\alpha$.}
\label{fig:supp_4}
\
\end{figure}

In the toy experiment of sequential task, We observe that neurons have quite similar activations, which turns the analysis of resources to the analysis of neuron redundancy. We measure how much redundancy lies in the model by quantifying the correlation between neuron activations within layers. To be specific, for each neuron, we count the ``neuronal redundancy", analogous to neuroscience, a number of other neurons in a layer that has highly correlated activation patterns ($\text{corr} > 0.75$). In Figure 4-B, we observe that a higher fraction of neurons in the hidden layer exhibit non-zero neuronal redundancy as $\alpha$ increases. The increase of redundant neurons suggests that the network gains redundancy overall regardless of specific functions each neuron serves in performing a composite function regression. 

The observed patterns of neuron redundancy imply two distinct regimes in how the model adapts to the task with the addition of neurons. In the first regime (where $\alpha < 640$), the model assigns novel functions to newly acquired neurons, leading to changes in the module's structure. Conversely, in the second regime (where $\alpha > 640$), the addition of neurons enhances previously existing functions, resulting in an increase in the module's size while its structure remains unchanged. Our resource model is predicting -1 exponent only when module is increasing in size, not their structure. Such conceptual picture can explain two different regimes of scaling observed in Figure ~\ref{fig4}-C.

\subsection{Seperability of sequential task loss}

We propose that the nature of the composite task loss, which is linearly separable into the sum of individual subtasks loss, accounts for observed scaling. Consider a model tasked to calculate a composite function $f \circ g$. Let us assume that the model performs this calculation in a composite manner, such that it first performs $\hat{g}: x \to \hat{g}(x)$ with $modules$ of $g$ and then $\hat{f}: x \to \hat{f}(x)$ with $modules$ of $f$. The Mean Squared Error (MSE) for the regression task for $f \circ g$ can be expressed as follows:

\begin{align*}
l &= \left\|f(g(x)) -\hat{f}(\hat{g}(x))\right\|^2 \\
&= \left\|f(g(x)) - f(\hat{g}(x)) + f(\hat{g}(x)) - \hat{f}(\hat{g}(x))\right\|^2 \\
&= \left\|f(g(x)) - f(\hat{g}(x))\right\|^2 + \left\|f(\hat{g}(x)) - \hat{f}(\hat{g}(x))\right\|^2 \\
&\quad + 2 \left\|\left(f(g(x)) - f(\hat{g}(x))\right)(f(\hat{g}(x)) - \hat{f}(\hat{g}(x)))\right\|^2
\end{align*}

Once we assume that each error function of $\hat{f}$ and $\hat{g}$, such that $e_f(x) = \hat{f}(x) - f(x)$ and $e_g(x) = \hat{g}(x) - g(x)$, are independent and orthogonal for the given input distribution, the last term vanishes. Meanwhile, the first term $ \left\|f(g(x)) - f(\hat{g}(x))\right\|^2 =\left\| f'(g(x)^*) e_g(x)\right\|^2$ (by Taylor series) represents the regression error of the $g$ part. Assuming $f'(g(x)^*)$ and $e_g(x)$ is independent and orthogonal, the first term is proportional to the $\left\|e_g(x)\right\|^2$. Similarly, the second term $ \left\|f(\hat{g}(x)) - \hat{f}(\hat{g}(x))\right\|^2$ represents the regression error of the $f$ part and is proportional to $\left\|e_f(x)\right\|^2$ under few assuptions. In summary, the task loss can be seperated to linear sum of MSE of $e_g(x)$ and MSE of $e_f(x)$ as follows:

\begin{equation}
l =  A \* \left\|e_g(x)\right\|^2 + B \* \left\|e_f(x)\right\|^2,
\end{equation}

The above derivation of linear seperability can be applied to any number of subtasks combined in series.

\subsection{Allocated Neurons by layers in the sequential task}

\begin{figure}[h]
\begin{center}
\includegraphics[width=8cm]{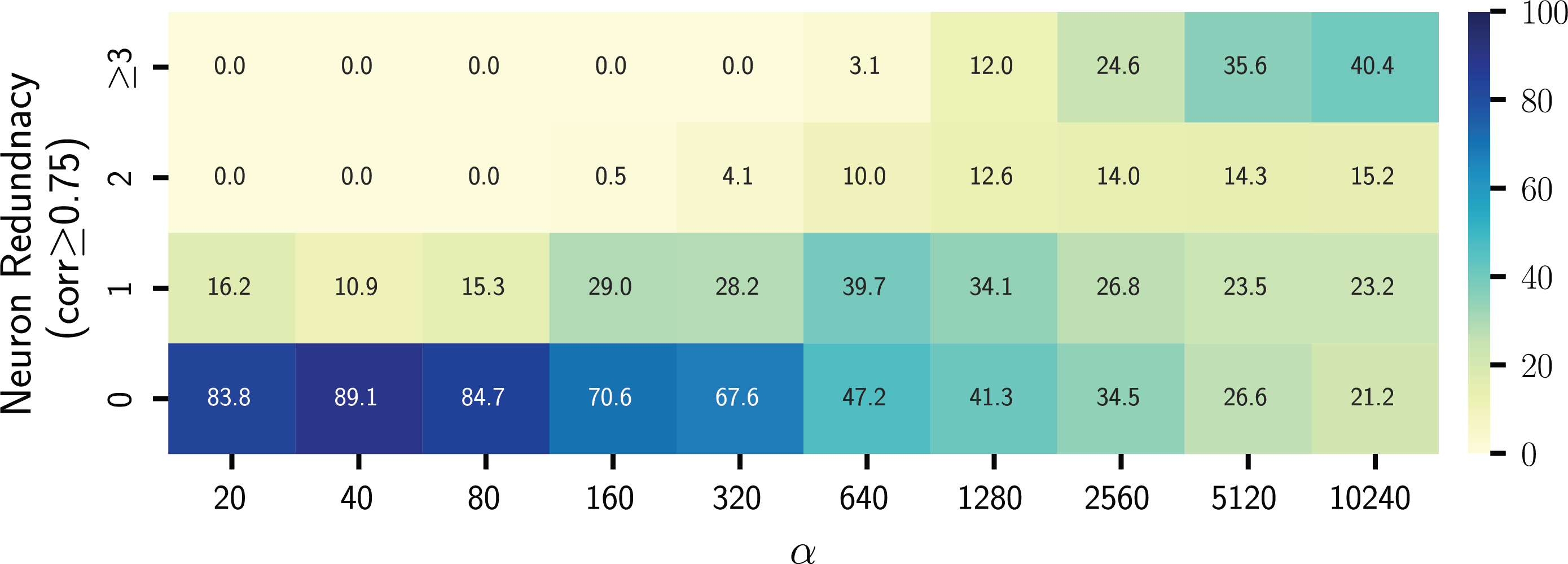}
\end{center}
\caption{Number of allocated neurons in each layer are plotted as respect to $\alpha$ in toy experiment for composition of tasks in series.}
\label{fig:supp_3}
\
\end{figure}

\end{document}

%% file: math_commands.tex
\usepackage{amsmath,amsfonts,bm}

\def\eqref#1{equation~\ref{#1}}

\def\1{\bm{1}}

\DeclareMathAlphabet{\mathsfit}{\encodingdefault}{\sfdefault}{m}{sl}
\SetMathAlphabet{\mathsfit}{bold}{\encodingdefault}{\sfdefault}{bx}{n}

